# Forecasting the Colorado River Discharge Using an Artificial Neural Network (ANN) Approach

Amirhossein Mehrkesh, Maryam Ahmadi
University of Colorado Denver, Denver, CO 80005

**Abstract**

Artificial Neural Network (ANN) based model is a computational approach commonly used for modeling the complex relationships between input and output parameters. Prediction of the flow rate of a river is a requisite for any successful water resource management and river basin planning. In the current survey, the effectiveness of an Artificial Neural Network was examined to predict the Colorado River discharge. In this modeling process, an ANN model was used to relate the discharge of the Colorado River to such parameters as the amount of precipitation, ambient temperature and snowpack level at a specific time of the year. The model was able to precisely study the impact of climatic parameters on the flow rate of the Colorado River.

1. Introduction

The volumetric flow rate of a river, also called its discharge, at a particular point, is the volume of water passing through the cross section of the river at that point in a unit of time. As aforementioned, forecasting the flow rate of a river could be very useful in water resources management. Any seasonal river basin planning for designation of water between different consumers can not succeed without knowing/predicting the amount of water (i.e. flow rate) passes through the river at that time.

## 1.1. Colorado River

The Colorado River is one of the main surface water streams in the southwestern United States. The 1,450 mile river flows from the beautiful Rocky Mountains in State of Colorado to the Gulf of California, draining 246,000 square miles of land area within the seven U.S. states. The Colorado River is a vital water supplier to the agricultural and municipal use, as well as the hydroelectric production in the southwestern U.S. [1]

The Colorado River originates at "La Poudre Pass" which is located east of "Never Summer Mountains" in Rockies of Colorado, some 60 miles northwest of Denver. The river flows towards the south of Colorado then it turns west after passing the Grand Lake [2].

"La Poudre Pass" lake is an elevated water reservoir, located in the Rocky Mountains and is notable for being the recognized source of the Colorado River depicted in Figure1 as point A. The upper basin of the Colorado River from the origin to the first discharge station, Baker Gulch, is also shown in the same figure.

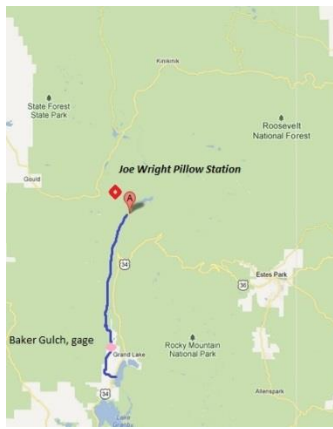

**Figure 1:** Colorado River course from the origin to the Baker Gulch gage

Colorado River in its natural state, pours around 15.7 million acre-feet into the Gulf of California each year, which is equivalent to an average flow rate of 21,700 cubic feet per second. The river discharge has been significantly varying based upon climatic conditions. The USGS web site has tabulated the average amount of monthly discharge of the Colorado River for different locations throughout the river course. [3,5]

**1.2. Artificial Neural Networks (ANNs)**

An artificial neural network (ANN) is a mathematical/computational model inspired by the structure of biological neural networks [4]. A neural network consists of an interconnected group of artificial neurons, which further processes the information through a connectionist approach to computation. Generally, an ANN model is an adaptive tool which is able to change its own structure based on the external or internal information flowing through during the learning step, and is used to model a complex relationship between input and output parameters. Inputs and output parameters required for creating an ANN model should be chosen based on these facts that the input parameters should be independent from each other and the output variables being a function of the inputs.

An ANN model usually consists of an input layer including the input parameters, and an output layer including the output (desired) parameters as well as one or more hidden layer/s, which latter is meant to make a connection between input and output parameters, thereby making the modeling plausible. In an ANN each layer has a number of nodes; for the input and output layers, the number of nodes is equal to the number of parameters, whereas being a model adjustable parameter for hidden layers. Changing the number of nodes in the hidden layer/s can help in improving the accuracy of the model to reach the best solution. A typical scheme of an ANN model is shown in Figure 2.

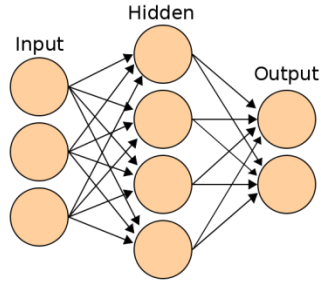

**Figure 2**: Typical scheme of an ANN

## 2. Material and methods

### 2.1. Database and assumptions

In order to study the impact of parameters on the discharge of a river, several climatic parameters have been previously considered important which are listed as follows.

1) Snowpack, Swe (snow water equivalent), in Colorado River origin measured at a specific time of the year here on May $1^{st}$.

2) Precipitation amount during the period in which snow melts.

3) The average temperature of ambient in snow melting period.

The relative humidity (RH%) and wind velocity could also influence the river discharge, but due to the lack of historical data from the weather stations throughout the region, these parameters were not taken into account.

In order to investigate the influence of climatic data (snowpack, precipitation and ambient temperature) on the discharge of the Colorado River, the upper part of the river from the origin, La Poudre Pass lake, to the first discharge station, Baker Gulch, was considered for the modeling purposes. The Baker Gulch station is a discharge measuring station located 5.5 miles northwest of the town of Grand Lake and its discharge data were appropriate for our study goal since there was no notable human-made diversions before that. It also should be mentioned that there are several transmountain diversions upstream of the discharge station close to the "Grand River" ditch, therefore no significant mountain runoff occurs prior to the station.

Snowpack data were extracted from the Snotel database, where the "Joe Wright Pillow" station found to be the closest station to the origin of the Colorado River, which its location is depicted in Figure1. The density of snow varies from year to year, therefore the snowfall amount cannot be considered as a parameter, thus the water equivalent amount of the snowpack (swe) on a particular date was considered as an input parameter. Different databases report the amount of snowpack on May $1^{st}$ of each year, which can be an appropriate parameter for predicting the average discharge of the river in the summer time, in addition, there is going to be an insignificant amount of snowfall after May $1^{st}$, therefore the amount of snowpack is not subject to a considerable increase, moreover the warmer weather after May $1^{st}$ begins to melt the snow. Studies show that the snowpack in the upper basin of the Colorado River vanishes by the end of July of each year as shown in Figure3.

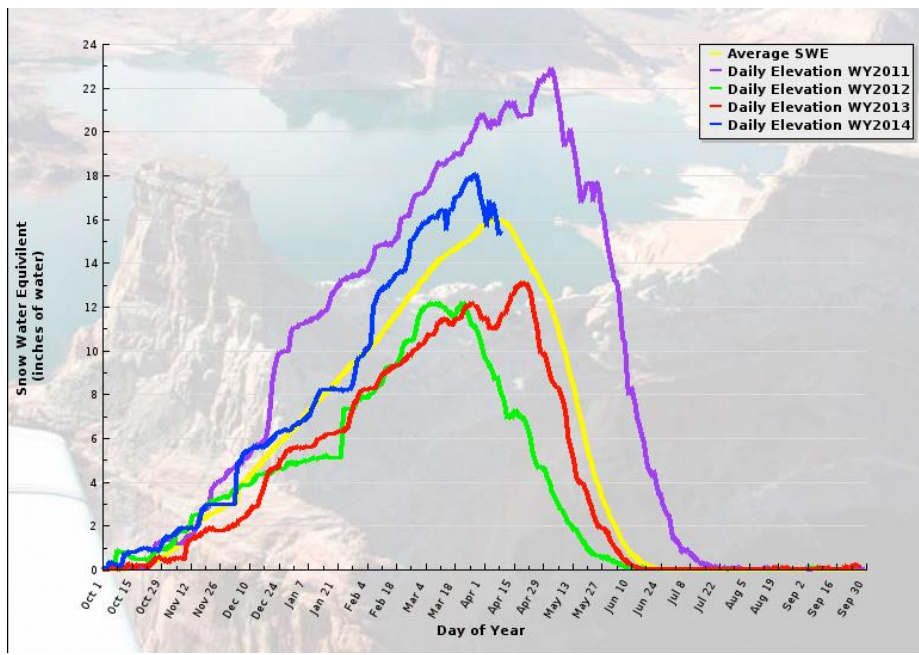

**Figure 3:** Snowpack fluctuations in the upper Basin of Colorado River [5]

It can be observed from Figure3 that the entire amount of the snow melts by the end of July, therefore the average amount of the discharge in May, June and July were considered as the output parameters of the ANN model. The amount of river discharge in these three months was collected from the USGS web site dataset for the Baker Gulch station. The average ambient temperature in these three months was chosen as one of ANN input parameters while the precipitation amount after May 1$^{st}$ was considered to be the second input parameter. The amount of precipitation in three chosen months was gathered from the Joe Wright Pillow's weather station database, then the cumulative summation of the precipitation in May, June, July in different years were selected to be entered into the ANN model. [6, 7, 8]

**2.2. Neural Network Architecture**

A feed-forward Multilayer Perceptron Neural Network, MLPNN, was trained using an adequate number of inputs and outputs from 1979 trough 2009, as tabulated in table1.

**Table1:** ANN input and output elements

| Input | Output |
|---|---|
| Snowpack (Swe) [inch] May 1$^{st}$ | Discharge (Avg. May, June, July) [CFs] |
| Precipitation [inch] (May, June, July) | |
| Mean Ambient Temperature [$^{0}$R] (Avg. May, June, July) | |

For modeling purposes, several networks with different architecture were trained and the most accurate network was achieved with the following characteristics.

- ➢ Number of hidden layers: 1 [9]
- ➢ Number of nodes in the Hidden layer: 7
- ➢ Hidden layer and Output layer transform function: Tanh
- ➢ Learning rule: Momentum & Updating: Batch

In order to compare the accuracy of the network having different number of hidden layer's nodes, three different parameters, MSE (Mean Square Error), r (Correlation coefficient between a calculated output and actual output) and Error%, were used to analyze the performance of the network when a particular set of data used.

The amount of these three values for different networks with various numbers of nodes in the hidden layer are listed in Table2.

**Table 2:** The amount of MSE, r, and % Error for different number of hidden layer's nodes

| Node | MSE | r | % Error |
|---|---|---|---|
| 3 | 0.000387 | 0.9941 | 2.0231 |
| 4 | 0.000368 | 0.9957 | 1.7342 |
| 5 | 0.000354 | 0.9968 | 1.6245 |
| 6 | 0.000321 | 0.9979 | 1.5262 |
| **7** | **0.000272** | **0.9993** | **1.3222** |
| 8 | 0.000298 | 0.9985 | 1.4356 |

Actual and predicted data for the discharge of Colorado River from 1979 to 2009 are shown in Figure4, where a good accordance with low error is observed.

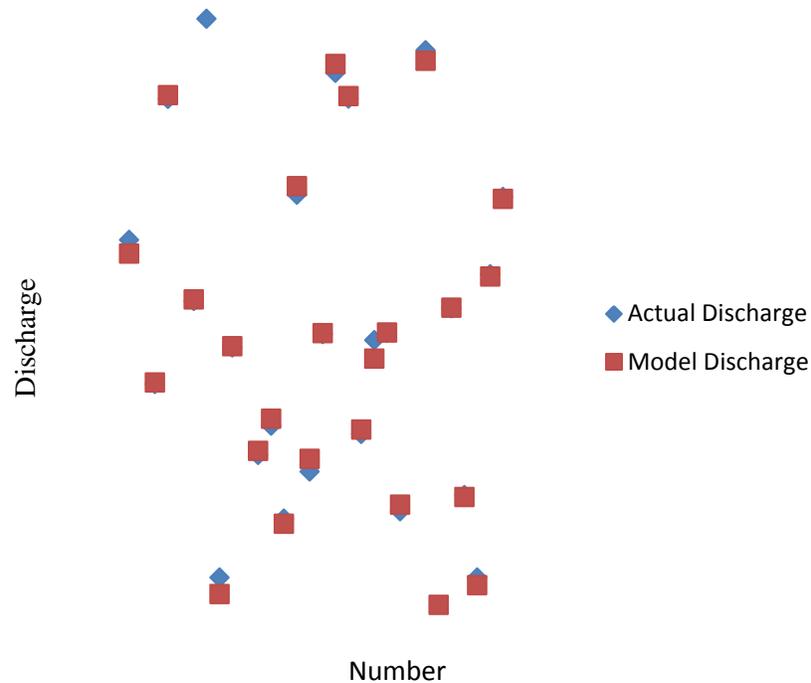

**Figure 4:** Actual values vs. calculated values of Colorado River discharge

## 2.3. Sensitivity Analysis

Data production is one of the useful abilities of an accurate ANN model, which can be utilized to optimize an objective function or to do a sensitivity analysis and a set of output data can be produced by any trained network. In order to perform a sensitivity analysis, two of these three input parameters kept fixed while the third one was changing within a certain range. In the next step, a dataset was entered into the trained ANN so that the corresponding outputs, the flow rate, was predicted. The sensitivity analysis performed in this paper showed us that how output parameters are influenced from the input parameters. A change in the amount of input parameters, may significantly affect the output parameter. The effect of change in the snowpack on the river discharge for different amount of precipitations and constant ambient temperature is shown in table 3.

**Table 3:** Discharge variations vs. change in snowpack for different precipitations at T= 45 $^0$F

| Precipitation = 4 in T= 505 R | | Precipitation = 8 in T= 505 R | | Precipitation = 12 in T= 505 R | | Precipitation = 16 in T= 505 R | |
|---|---|---|---|---|---|---|---|
| Snowpack Change* | Discharge change | Snowpack Change | Discharge change | Snowpack Change | Discharge change | Snowpack Change | Discharge change |
| 2 | 1.983 | 2 | 1.963 | 2 | 1.944 | 2 | 1.926 |
| 3 | 2.967 | 3 | 2.928 | 3 | 2.890 | 3 | 2.854 |
| 4 | 3.952 | 4 | 3.893 | 4 | 3.837 | 4 | 3.782 |
| 5 | 4.938 | 5 | 4.859 | 5 | 4.784 | 5 | 4.711 |

* The snowpack change values mean that the basis amount was multiplied by that number, e.g. snowpack change of 2 means that the snowpack amount was considered to be doubled and so on.

The effect of change in precipitation on the discharge of the river for various amounts of snowpack at a constant temperature is presented in Table4.

**Table 4:** Discharge variations vs. change in precipitation for different snowpack at T= 45 $^0$F

| Snowpack = 5 in T= 505 R | | Snowpack = 10 in T= 505 R | | Snowpack = 15 in T= 505 R | | Snowpack = 20 in T= 505 R | |
|---|---|---|---|---|---|---|---|
| Percip. Change | Discharge Change | Percip. Change | Discharge Change | Percip. Change | Discharge Change | Percip. Change | Discharge Change |
| 2 | 1.0203 | 2 | 1.0103 | 2 | 1.0069 | 2 | 1.0051 |
| 3 | 1.0407 | 3 | 1.0205 | 3 | 1.0137 | 3 | 1.0103 |
| 4 | 1.0610 | 4 | 1.0308 | 4 | 1.0206 | 4 | 1.0154 |
| 5 | 1.0814 | 5 | 1.0410 | 5 | 1.0274 | 5 | 1.0206 |

The effect of change in the ambient temperature on the discharge at a fixed precipitation rate and snowpack is translated into the data listed in table5.

**Table 5:** Discharge change *vs.* temperature difference at precipitation = 4 in & Snowpack = 5 in

| Snowpack= 5 in, Precipitation= 4 in  Temperature range in the database = 505 R, 510 R (45 – 50 $^0$F) ||
|---|---|
| **Temperature variation (F)** | **Discharge Decrease (%)** |
| +2.5 | 2.169192 |
| +5 | 4.405687 |

The amount of the river discharge decreases when the ambient temperature increased.

## 3. Conclusion

- ANN could be safely used for modeling the river discharge, based on different climatic parameters such as snowpack, precipitation, and the ambient temperature.
- The ANN model was accurate enough to be used for a sensitivity analysis.
- The Snowpack amount considered to be the most effective parameter and the ambient temperature has the least effect on the river discharge.
- The average amount of the river discharge increases when snowpack and precipitation rate both increases and decreases when ambient temperature goes higher.

**References**

1. Diaz, Henry F.; Anderson, Craig A. (2003), "*Precipitation Trends and Water Consumption in the Southwestern United States*". Impact of Climate Change and Land Use in the Southwestern United States, U.S. Geological Survey.

2. Lindberg, Eric (2009), "*Colorado off the Beaten Path: A Guide to Unique Places*". Globe Pequot, ISBN: 0-7627-5024-3, P.38.

3. Hoover Dam 75$^{th}$ Anniversary History Symposium, pp. 101–102

4. J. J. Hopfield, (1982), "*Neural networks and physical systems with emergent collective computational abilities*". Proc. Natl. Acad. Sci. USA, Vol. 79, pp. 2554-2558.

5. http://graphs.water-data.com/ucsnowpack/

6. www.usgs.gov

7. http://www.wcc.nrcs.usda.gov/nwcc

8. http://weather-warehouse.com/WeatherHistory

9. G. Cybenko, (1989), "*Approximations by superpositions of a sigmoidal function Math*". Cont. Signal Syst. 2, pp. 303-314.